# Learning for Biomedical Information Extraction: Methodological Review of Recent Advances

Feifan Liu[1], Jinying Chen, Abhyuday Jagannatha, Hong Yu

## Abstract

Biomedical information extraction (BioIE) is important to many applications, including clinical decision support, integrative biology, and pharmacovigilance, and therefore it has been an active research. Unlike existing reviews covering a holistic view on BioIE, this review focuses on mainly recent advances in learning based approaches, by systematically summarizing them into different aspects of methodological development. In addition, we dive into open information extraction and deep learning, two emerging and influential techniques and envision next generation of BioIE.

## Introduction

Biomedical information extraction (BioIE) aims to automatically unlock structured semantics (e.g. entities, relations and events) out of unstructured biomedical text data. It has been successfully applied in clinical decision support [1,2], integrative biology [3,4] and biocuration assistance [5–7], and has also shown great potentials in pharmacovigilance [8].

Broadly speaking, BioIE covers a very large spectrum of research efforts, resulting in enormous publications, as shown in Figure 1. The tasks include named entity recognition [9–11], event identification[12–14], and relation extraction [10,15,16]. The text data or corpora include medical

[1] Corresponding author. E-mail: feifan.liu@gmail.com

**Feifan Liu**, PhD, is Senior Research Scientist in the Clinical Language Understanding group, Healthcare division at Nuance Communications. His research interests include natural language processing, machine learning and biomedical informatics.

**Jinying Chen**, PhD, is Research Scientist in the BioNLP Lab at the Department of Quantitative Health Sciences of University of Massachusetts Medical School. Her research interests include natural language processing, machine learning and their applications in biomedical information extraction.

**Abhyuday Jagannatha** is PhD student at the School of Computer Science, University of Massachusetts, Amherst. He is a research assistant in the BioNLP Lab at UMass Medical School, under the supervision of Prof. Hong Yu. The main focus of his research is natural language processing and its applications in health informatics.

**Hong Yu**, MS, MA, PhD, FACMI, is Professor at the Department of Quantitative Health Sciences of University of Massachusetts Medical School and Research Health Scientist at VA Central West Massachusetts.

literature[17], biological literature[18], electronic health records[19], and healthcare related social media data[20]. The methodology includes rule-based method, knowledge-based method, statistics based method, learning-based method and hybrid method[21–23]. A thorough systematic survey covering all related work in BioIE is beyond the scope of this paper, and readers of interest can refer to several previously published survey papers[24,25,19,26,27,22,28–30,21,23,31,32]. Most recently, Gonzalez et al.[33] summarized text and data mining advances and emerging applications for biological discovery, which is domain specific and task oriented, focusing on applications of natural language processing (NLP) techniques to better understand underlying mechanisms of disease. It is IE related, but mainly for knowledge discovery. Instead this review will focus on technological advances for learning-based BioIE across different biological, medical and clinical domains, which will yet be abstracted across different types of tasks, different genres of corpora, as well as different subdomains being applied, shedding light on future research directions and providing prospective insights into next generation of BioIE.

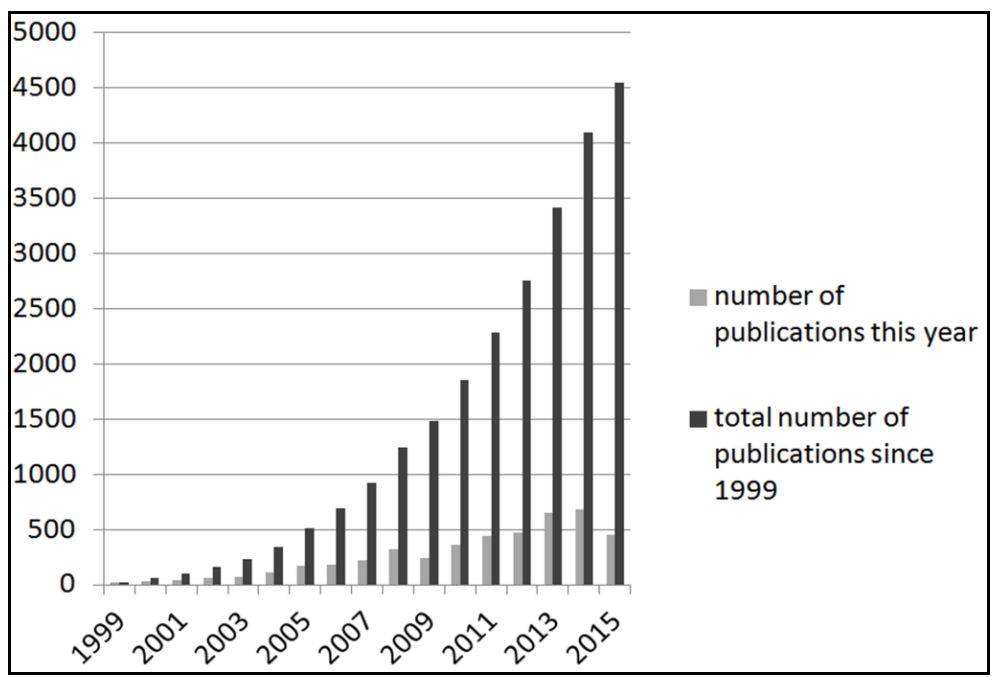

Figure 1. The number of publications in biomedical information extraction each year (grey bar) and cumulative number since 1999 (dark bar), using the PubMed query of "(biomedical OR clinical OR medical) AND (text+mining OR text+processing OR natural+language+processing) AND (information+extraction OR named+entity+detection OR named+entity+recognition OR relation+extraction OR event+extraction)".

## Learning for Biomedical Information Extraction

Learning methods for BioIE have been rapidly advanced in the past 5 years. Conditional random fields(CRF) [10] and structured support vector machines(SSVM) [34] have been two representative learning methods, and deep neural networks[35] have been increasingly applied to both general domain IE and BioIE. All those methodological advances mainly fall into three categories: (1) learning from labeled data (i.e. supervised learning) which focuses on labeling more data to tackle new problems or improving model training using existing benchmark data; (2) learning from unlabeled data (i.e. semi-supervised and unsupervised learning) which involves incorporating large amount of unlabeled data into the learning process; (3) learning scheme integration to seamlessly integrate different learning paradigms at outer system level (i.e. hybrid approach) or modeling level (i.e. joint modeling). In particular, two emerging technological advances, namely open information extraction (OpenIE) and deep learning (DL), have exerted significant influence on BioIE recently for more scalable and reliable applications. What follows will be reviewing each of the abovementioned topics in more details.

**Learning from Labeled Data**

Traditional learning-based systems aim to infer optimal prediction functions from labelled training data instances, which can be used for mapping new data instances to their (predicted) labels. There are mainly two types of approaches that facilitate pattern analysis on labeled data: (1) feature-based approach explicitly transforms raw data representation into feature vector space, and each feature dimension represents an individual characteristics of a data instance; (2) kernel-based approach provides an implicit way of transforming raw data into a high-dimensional space through a similarity function, called kernel function, which is defined over pairs of data instances.

- *Feature-based Approaches*

Advanced feature engineering has proven to be successful in many machine learning based BioIE applications. In addition to common features, such as lexical (e.g., words), orthographic (e.g., capitalization, character *n*-grams), syntactic (e.g., part-of-speech, chunking), semantic (e.g., semantic category of a word by dictionary or ontology) and local context (e.g., *n*-grams, neighboring words) features [36–39], discourse-level features (e.g., sentence length, section headings, writing patterns) [36,37] and dependency tree based features [39–42] have also been explored. Different systems also leverage task specific features using external resources or other domain knowledge, e.g. chemical elements, amino acid sequences and chemical formulas were used to derive features for recognizing chemical named entities[43]; features derived from event occurrence pattern were designed to detect the causal relation in a clinical problem-action relation extraction system[44]. To improve the performance of feature-based system, combining multiple types of features is a widely used strategy, e.g. a drug-drug interaction (DDI) system[45] was developed to combine heterogeneous features, including lexical, syntactic, semantic and negation features derived from sentences and their corresponding parse trees.

Word features are prone to the data sparseness challenge (i.e., insufficient training data vs. a huge amount of features). To overcome such challenge, vector-based word representations have been exploited, including clustering-based word representation (i.e., representing a word by its hard cluster(s)) [36,46], distributional representation (i.e., representing a word by its semantically related words as calculated on word co-occurrence statistics) [37,46], and distributed representation (i.e., word embedding) [46,47]. Tang et al. [48] combined clustering-based word representation and distributional word representation into a structural SVM learning scheme, showing better performance than using either single type of representation feature. More on word embedding will be described later in deep learning section.

However, a richer set of features may not guarantee better performance, as features that are irrelevant, redundant or have limited discriminative power may cause adverse effects on model learning, leading to increased computational complexity and over-fitting. As a result, feature selection is of paramount importance for any learning-based approaches, especially with high-dimensional features.

Simple additive or subtractive feature selection strategies (adding or removing each feature class one by one to evaluate feature contribution) work well for many BioIE applications (e.g. [49]). New feature selection approaches were also proposed, e.g. Leaman [50] applied multivariate feature selection with filters and false discovery rate analysis to biomedical NER; Xia et al. [51] evaluated an accumulated effect evaluation (AEE) algorithm for feature selection and showed improved performance in the GENIA bioNLP shared task 1 (event detection); Campos et al. [52] described an optimization algorithm to find the feature set that better reflect the characteristics of each event type for biomedical event trigger recognition; and Fang et al. [53] introduced an improved feature selection method by combining mutual information and class separability criteria for identifying non-redundant optimal features on multidimensional time series clinical data. In addition, a partition-specific feature selection method was successfully applied on both protein-protein interaction[54] and drug-drug interaction [45] tasks, where candidate relation pairs are partitioned into groups based on syntactic properties and features are optimized for each group, achieving the state-of-the-art system performance.

- *Kernel Approaches*

Kernel-based approaches are becoming more and more popular for learning algorithms including perceptron and support vector machines, as sometimes data can't be easily represented with explicit feature vectors (e.g. sentences are better characterized by trees or graphs). In the last few years, many task specific kernels have been proposed or applied on many BioIE tasks. Patra et al. [55] proposed a novel kernel named "sliding tree kernel", which is an improved tree kernel specific to named entity recognition(NER) task. It considers a substructure of trees in the form of sliding window, leading to a better way to model local context of a word. A hash subgraph pairwise(HSP) kernel was introduced for protein-protein interaction extraction[56], which can efficiently make use of the full dependency graph that represents sentence structure and particularly capture the contiguous topological and label information.

In addition to single-kernel-based approaches, ensemble approach combines outputs from different kernel-based systems and multiple kernel learning (MKL) approach constructs a hybrid kernel by linearly or polynomially combining individual kernels. Thomas et al. [57] reported the best performing drug-drug interaction system in the DDI extraction 2011 challenge, which combined outputs from two kernel-based systems and a case-based reasoning system using majority voting. Yang et al. [58] presented a weighted MKL approach for protein-protein interaction(PPI) from biomedical literature, where different kernels (feature-based, tree, graph and part-of-speech path) were combined linearly and the shortest path-enclosed tree(SPT) and dependency path tree were extended to capture richer contextual information, achieving better performance than counterpart systems. Following this strategy, Li et al. [59] further improved the state-of-the-art performance on PPI by incorporating a semantic kernel characterizing the protein-protein pair similarity using Medical Subject Heading and the context similarity using WordNet. Similarly, the best system [60] in SemEval'13 DDI extraction applied a MKL approach linearly combining a feature-based kernel, a shallow linguistic kernel and a path-enclosed tree kernel. Another hybrid kernel, which consists of an entity kernel and convolution parse tree kernel combined via polynomial expansion, was showing promising result on biomedical event extraction[61].

Tikk et al. [62] conducted an analytical study on the performance of 13 types of kernels for PPI extraction, which suggests that the system performance benefits more from novel features than from novel kernel functions. Rich features have been explored to improve kernel-based systems. For instance, Zhou et al. [63] showed a novel framework for biomedical event trigger identification, where word embedding features were combined with syntactic and semantic contextual features using MKL method, achieving the state-of-the-art performance. Similarly, Li et al. [64] showed improved performance when integrating word embedding features in a kernel based PPI extraction system, and Ma et al.[65] improved PPI extraction by proposing a new tree kernel where processing rules were defined to better handle the parsing error of modal verb phrases and noise interference by appositive dependency. In contrast, Kim et al. [66] showed that a simple linear kernel that integrates a rich set of lexical and syntactic features is able to achieve a competitive performance for DDIs, suggesting that linear kernel may perform as well as complex kernels.

- *Cost-effective Ground Truth Acquisition*

Supervised machine learning approaches depend on annotated corpora, which are frequently expensive to obtain, especially in the medical/clinical domain. Semi-supervised annotation, active learning, and crowd-sourcing approaches have been developed to create large-scale annotated corpora.

Pre-annotation or computer-aided annotation gives human annotators the machine-annotated data for potentially better efficiency. Recent studies, however, reported mixed results in terms of how much pre-annotation helps. Lingren et al.[67] created a dictionary for pre-annotation on clinical NER task, showing reduced time needed for review by 13.85-21.5% compared with fully manual annotation. However, in an experiment on clinical text de-identification task by South et al.[68], it has been shown that machine-assisted annotation didn't improve annotation quality for most PHI classes and didn't provide statistically significant time-savings compared to manual annotation of raw documents. By further combining iterative machine learning approach in pre-annotation, RapTAT[69], a semi-automated semantic annotation tool, was shown capable of reducing the annotation time by up to 50% on 401 clinical notes, as well as improving the inter annotator agreement.

Active learning aims to alleviate annotation efforts by reducing annotated sample size. It selects informative samples via actively involving the learning algorithms. Various studies have investigated the effects of using active learning to achieve less learning cost without compromising learning performance of associated predicative models, such as assertion annotation for medical problems[70], semantic annotation for medical abbreviations and terms[71], clinical NER annotation[72], clinical coreference resolution[73], pathological phenomena labeling in MEDLINE[74] and phenotype annotation[75]. Two recent studies on medical concept extraction show that: (1) incorporating external knowledge resources within active learning query strategies[76] can further reduce annotation efforts; (2) incremental active learning[77] is promising in building robust medical concept extraction models while significantly reducing the burden of manual annotation.

Crowdsourcing has been widely explored in biomedical and clinical domain [78]. Studies have demonstrated that crowdsourcing is an inexpensive, fast and practical approach for collecting high-quality annotations for different BioIE tasks, e.g. medication NER in clinical trial documents[79], disease mention annotation in PUBMED literature[80], relation extraction between clinical problems and medications[81], etc. Different techniques have been explored to improve the quality and effectiveness of crowdsourcing, including probabilistic reasoning[82] to make sensible decisions on annotation candidates and gamified techniques[83] to motivate a continuous involvement of the expert crowd. More recently, a method called CrowdTruth [84] was proposed for collecting medical ground truth through crowdsourcing, based on the observation that disagreement analysis on crowd annotations can compensate lack of medical expertise of the crowd. Experiments of using CrowdTruth on medical relation extraction task show that the crowd performs just as well as medical experts in terms of quality and efficacy of annotation, and also indicate that at least 10 workers per sentence are needed to get the highest quality annotation for this task.

**Learning from Unlabeled Data**

In contrast to relying on the costly labeled data, freely available unlabeled data have been explored for BioIE. Approaches include unsupervised, semi-supervised, and distant supervision.

Unsupervised biomedical NER systems described in [85] are based on phrase chunking and distributional semantics, showing competitive results on both clinical notes and biomedical literature. Quan et al. [86] explored kernel-based pattern clustering and sentence parsing to tackle the PPI extraction and gene-suicide association extraction from biomedical literature. More recently, Alicante et al. [87] reported an unsupervised system for both entity and relation extraction from clinical records written in Italian.

Semi-supervised methods aim to incorporate unlabeled data in a supervised manner. Recent semi-supervised approaches in BioIE mostly differ in the approximation methods used to obtain labelling for unlabeled data as well as the ways to handle uncertainty when adding unlabeled data, including self-training, also known as bootstrapping (e.g. [88] for medical risk event extraction and [89] for drug-gene relation extraction), transfer learning (e.g. [90] for clinical concept extraction), and manifold regularization (e.g. [91] for medical relation extraction). To obtain higher quality labelling for unlabeled data, several strategies have been proposed for semi-supervised learning, such as combining active learning for PPI extraction[92], introducing event inference mechanisms to detect more event mentions from unlabeled text [93], and exploiting topic analysis to identify similar sentences for automatic labeling[94]. From feature representation point of view, feature coupling generalization(FCG) has been explored to generate new features directly from unlabeled data, e.g. [95] for gene NER and PPI extraction and [96] for biomedical event extraction.

Different from abovementioned semi-supervised methods, distant supervision approach typically doesn't require any labeled data, which makes use of weakly labeled data derived from a knowledge base[97], or some seed data points[98] but without finer-grained sentence level annotations. Recently it has been widely used in DDI and PPI extraction[97,99,100] with different strategies to deal with training challenges due to noisy data, as well as later work on the BioNLP Gene Regulation Network task[98], gene expression relation extraction[101] and cancer pathway extraction[102]. Most recently, the DeepDive system[103] was employed to scaled up the distant supervision method on biomedical text mining without limiting the application to a specific process or pathway, achieving promising results for extracting gene interactions from full-text PLOS articles.

**Learning Scheme Integration**

- *Hybrid Approach*

Hybrid approaches integrate heuristics/rule/pattern-based method, domain knowledge, and learning-based method. One ensemble strategy is to develop multiple independent models, and then combine the results of each model for a final output, either through rules or by using some classification/regression model, e.g. combining rule-based model with SVM classifiers for biomedical event detection[104]; integrating pattern recognition into learning for DDI extraction[105]; algorithmically fusing results from two approaches for temporal relation extraction in patient discharge summaries[106]. Another ensemble strategy is to run different models sequentially for further filtering and refining to a better system output, e.g. extracting disorder mentions from clinical notes [107]; extracting disease-treatment relations from MEDLINE corpus[108]; identifying Genia events by learning-based classifiers with rule-based post-processing [109]; and similar post-processing was also used in a hybrid system of recognizing composite biomedical named entities[110].

- *Joint Modeling*

A BioIE system typically involves different subtasks with embedding and inter-dependent characteristics. To overcome cascading errors in a multi-step pipeline framework, joint models (e.g. a Markov Logic Network(MLN) based approach [111]) have shown improved performance. But due to complexity of inference for joint modeling, rich features previously used in the pipeline framework may be compromised to make inference trackable. Therefore, many efforts have been made to relieve the computational bottleneck of joint inference in biomedical event extraction. For instance, Riedel and McCallum [112] proposed three joint models of increasing complexity to make the system more robust and applied dual decomposition to make joint inference tractable; Vlachos and Craven [113] applied the search-based structured prediction framework (SEARN) for high modeling flexibility and fast joint inference; Venugopal et al.[114] presented an MLN-based join model which employs an SVM model to encode high-dimensional features; and more recently Li et al.[115] exploited dual decomposition based joint inference, dependency parsing based rich features, and word embedding for event extraction. From a different perspective, Wei et al.[116] proposed to use dependency chain tagging to extract embedded semantic relations from biomedical literature, which avoided complex inference but kept the hierarchy feature of entities.

In addition to the abovementioned work focusing on BioNLP shared event extraction task, joint modeling has been increasingly introduced in other specific subdomains. To effectively extract adverse drug event(ADE), Li et al.[117] designed a transition-based model to extract drugs, disease and ADEs jointly, which leverages the structured perceptron for training and the multiple-beam search algorithm for decoding.

**Open Information Extraction (OpenIE)**

OpenIE [118] has been emerging as a novel information extraction paradigm in the last couple of years. It doesn't presuppose a predefined set of relations but aims to identify any possible relations from unlabeled data with no or limited supervision.

OpenIE systems typically consist of four main components: (1) Automatic Labeling of data using heuristics or distant supervision; (2) Extractor Learning using relation-independent features on noisy self-labeled data; (3) Tuple Extraction on a large amount of text by the Extractor; (4) Accuracy Assessing by assigning each tuple a probability or confidence score. Based on features used in extractor learning, we can roughly group existing OpenIE systems into two categories: Light Open Extractors (e.g. [118–121]) which only use shallow language processing, e.g. part-of-speech tagging and chunking, and Heavy Open Extractors (e.g. [122,123]) which use deep linguistic analysis, e.g. dependency parsing. Typically, the former is much more efficient but is much lower in either recall or precision, while the latter improves the overall performance significantly with the hit on system efficiency. OpenIE requires no or little supervision and its extractor is trained on automatically labelled data using heuristics or distant supervision, which is the same as traditional bootstrapping system in this regard. However, it lends itself well to contributing to next generation information extraction systems due to its particularly advantageous aspects, as summarized in Table 1.

Table 1. Advances of OpenIE Compared with Traditional self-learning

| OpenIE | Traditional self-learning |
|---|---|
| Highly scalable to size and diversity of the WEB | Relatively small and homogeneous corpus |
| Not dependent on relation specific features | Relation dependent features |
| Avoid lexical features for generalization | Use lexical features for better precision |
| Domain independent | Domain dependent |
| No predefined relation schema | Targeted to specific types of relations |
| Label all the data | Selectively label data |
| Redundancy-based accuracy assessing | Confidence derived from the learned model |

One major challenge is that the massive facts OpenIE systems extract are in purely textual surface form which cannot be directly used by applications. There are several directions to address this issue: (1) Building knowledge resources to make sense of OpenIE extractions, e.g. dynamic knowledge graphs[124], web-scale knowledge repository[125], taxonomy of textual relational patterns[126]; (2) Harnessing semantic web technologies to enable information fusion and semantic reasoning[127]; and (3) Integrating ontological resources, e.g., linking OpenIE to world knowledge[128,129] and aligning multiple ontologies [130]. In addition, Riedel et al. [131] proposed a universal schema, which unites surface form relational patterns from OpenIE with relations defined in knowledge bases. By using matrix factorization and collaborative filtering, this approach is able to reason about unstructured and structured data in mutually-supporting ways.

The great potential of applying OpenIE techniques to BioIE has been recognized. For instance, Attias et al. [132] adapted the Never Ending Language Learner (NELL) [133] to biomedical domain and proposed a Rank-and-Learn methodology to effectively prevent semantic drift, achieving promising results on learning biomedical lexicon of open categories. Nebot et al. [134,135] presented a scalable method that extracts surface-form biomedical relations without pre-specified types, and then infers abstract relations and their signature types by clustering these extracted relations.

**Deep Learning**

Deep learning refers to "a class of machine learning techniques that exploit many layers of non-linear information processing for supervised or unsupervised feature extraction and transformation, and for pattern analysis and classification"[136]. Deep learning networks can be roughly categorized into (1) unsupervised/generative, e.g., restricted Boltzmann machines (RBMs)[137], deep belief networks (DBNs)[138]; (2) supervised/discriminative, e.g., deep neural networks (DNNs)[139], convolutional neural networks (CNNs)[140] and recurrent neural networks(RNNs)[141]; and (3) hybrid, e.g., DBN-DNN[142] models that combine unsupervised pre-training and supervised fine-tuning.

There have been a surge of interest of applying deep learning techniques in common NLP tasks, such as semantic parsing[143], information retrieval[144,145], semantic role labeling[146,147], sentimental analysis[148], question answering[149–152], machine translation[153–157], text classification[158], summarization[159,160], text generation[161], as well as information extraction, including named entity recognition[162,163], relation extraction[164–168], event detection[169–171]. CNNs and RNNs are two popular models employed by these work. Despite of the difference in network architecture, they share the key motivation to avoid task-specific, knowledge-centric feature engineering by leveraging the word embedding technique.

More recently, remarkable progress has also been made in BioIE due to the widely propagated application of deep learning techniques. One popular technique from deep learning is word embedding, which have been widely used in biomedical named entity recognition[46,172], medical synonym extraction[173], medical semantics modeling[174], relation extraction including chemical-disease relation [175], Protein-Protein interaction[176], and relation between pharmaceuticals and diseases/physiological processes. In terms of event extraction, Li et al.[177] reported comparable state-of-the-art performance when applying word embeddings on BioNLP event extraction task; Nie et al.[178] presented an embedding assisted neural network prediction model to conduct biological event trigger identification; Henriksson et al.[179] leveraged distributed semantics to detect adverse event signals from clinical notes. Furthermore, Wu et al.[180] utilized neural word embeddings for disambiguating clinical abbreviation mentions; similarly Liu et al.[181] exploited task-oriented resources to learn word embeddings for clinical abbreviation expansion. Ghassemi et al.[182] employed distributed stochastic neighbor embedding (tSNE) to extract clinical sentiment information (positive vs. negative).

Instead of using standard word embedding training strategy, Jiang et al.[183] proposed a biomedical domain-specific word embedding model by incorporating stem, chunk, and entity information, and used them for DBN-based DDI extraction and RNN-based gene mention extraction. Similarly in clinical domain, Choi et al.[184] presented a novel neural word embedding tool, Med2Vec, which can not only learn distributed representations for both medical codes and visits in electronic health record (EHR), but also allow interpreting the learned representations confirmed positively by clinical experts. Then an extensive and comparative study[185] has been performed among medical concept embeddings learned from medical journals, medical claims and clinical narratives respectively, evaluating their similarity and relatedness properties.

Different deep learning architectures have also been explored to predict clinical events, e.g. Choi et al.[186] applied RNNs to longitudinal EHR data to predict future disease diagnosis and medication prescription. Following that, they further explored two gating mechanisms Long Short-term Memory (LSTM)[187] and Gated Recurrent Units (GRUs)[153] on RNN models, incorporating patient static data to extract early signs from EHR data for predicting kidney transplantation endpoint events, Jagannatha and Yu[188] employed a bidirectional LSTM RNN structure to extract adverse drug events from electronic health records, and Miotto et al.[189] explored a three-layer stack of denoising autoencoders to learn a general-purpose patient representation from EHR data, resulting in improved clinical predictions. In addition to using textual data, Liu et al.[190] proposed a recurrent self-evolving fuzzy neural network (RSEFNN) to leverage brain dynamics data to predict driving fatigue event; Lehman et al.[191] combined CNNs with dynamical systems to model physiological time series for prediction of patient prognostic status.

Due to the big success of deep learning on computer vision, information extraction/detection from medical imaging has been substantially influenced, such as pulmonary nodule detection in computed tomography scans using CNN features[192], a successful application of CNN trained on a large-scale non-medical image dataset for chest x-rays pathology detection[193], automated detection of posterior-element fractures on spine CT based on deep CNN [194], an interleaved text/image deep learning system for extracting and mining the semantic interactions of radiology images and reports[195]. More recently, Shin et al.[196] exploited and extensively evaluated three important factors of applying deep CNN to medical image classification, i.e., network architecture, dataset characteristics and transfer learning, in the context of two different computer-aided diagnosis applications: thoraco-abdominal lymph node detection and interstitial lung disease classification. Other successful examples of applying deep learning techniques include genomic information extraction[197–199], protein structure prediction[200] as well as drug discovery[201].

# Challenges and Future Directions

Recent innovated methodology development has significantly advanced biomedical information extraction, giving rise to a broad spectrum of improved biological and clinical applications. However, a number of limitations and problems at the frontiers of BioIE continue to impose additional challenges and present new opportunities for more accurate, efficient, reliable, scalable and sustainable BioIE research.

Data-driven approaches will continue to be a mainstream strategy. OpenIE techniques have been drawing more and more attention to enhance and scale BioIE systems by utilizing large, complex and heterogeneous data (different genres of textual data, structured vs. unstructured, text data vs. high throughput biological data) and extracting all meaningful relations and events without any restriction. Although in general domain several efforts have recently been made along this line, such as joint embedding of text and knowledge base (e.g. [202–205]) and unsupervised Web-scale event extraction[206], adapting current OpenIE techniques into BioIE applications is still at an early age. In addition, confidence based quality control and information normalization/fusion will be two remaining challenges for OpenIE platforms.

Joint inference models have been proposed to overcome the error propagation problem in pipeline approaches, by making predictions for multiple IE tasks simultaneously [207,208]. However, these models are complex and difficult to design. Exact inference is usually impossible and even inexact inference can be computationally expensive. Important questions remain to be answered in order for joint models to play a bigger role in BioIE, such as to what extent do we want a joint modeling? Which model architectures and learning methods better suit a specific IE problem? How to balance the computational efficiency and accuracy?

Evidently, deep learning techniques have played an unprecedented role in recent BioIE advances, which brought significant improvements across various subtasks. It is also signifying its continuous great potential for more advanced BioIE applications. Yet many issues remain to be investigated to take full advantage of deep learning for BioIE. Firstly, although deep learning can learn an internal distributed feature representation automatically from the data, combining domain knowledge and biomedical knowledge resources into deep learning architecture may lead to greater accuracy and flexibility. Collobert et al. [209] showed that by integrating linguistic and domain features, their DNNs improved over the original model (which does not use any feature engineering) and also outperformed benchmark systems in POS tagging and NER. The same strategy can be explored for BioIE. Secondly, deep learning models provide a natural way to learn distributed representations of entities and relations from knowledge bases [210,211]. Such approaches can be extended to fuse knowledge from multiple resources (e.g., knowledge bases, annotated corpora, and output from automatic systems). Finally, optimizing deep learning architecture to scale for big data processing and enable learning transferable features[212] will arouse increasing interest in the near future.

BioIE involves complex events, and those events are dependent. Better utilizing those inter-event dependencies[213] will be beneficial for further improving BioIE system performance, as well as for developing more integrative next-generation biomedical event extraction system through interconnecting biological reaction networks[214].

The big-data-driven BioIE research is rapidly evolving and projecting a bright future equipped with both learning-based algorithmic advances and multiple level model integrations.

> **Key Points**
>
> - Learning-based data-driven approaches have become and will continue to be the mainstream strategy for BioIE applications.
> - Methodological development evolves into joint modeling, system integration and embracing a variety of linguistic knowledge and domain-specific semantics.
> - Utilizing abundant unlabeled data presents promising results for tackling all kinds of BioIE challenges in a cost-efficient way, and will remain an active research area.
> - OpenIE and deep learning have shown significant potential in further advancing BioIE research, and are foreseen to be more involved for building cutting-edge BioIE solutions in the big data era.

## Funding

This work was supported by the National Institute of General Medical Sciences of the National Institutes of Health [grant number R01GM095476]. The content is solely the responsibility of the authors and does not necessarily represent the official views of the National Institutes of Health.

## References

1. Friedman C, Hripcsak G, Shagina L, et al. Representing information in patient reports using natural language processing and the extensible markup language. J Am Med Inform Assoc 1999; 6:76–87
2. Cao Y, Liu F, Simpson P, et al. AskHERMES: An online question answering system for complex clinical questions. Journal of biomedical informatics 2011; 44:277–288
3. Rebholz-Schuhmann D, Oellrich A, Hoehndorf R. Text-mining solutions for biomedical research: enabling integrative biology. Nature Reviews Genetics 2012; 13:829–839
4. Nikitin A, Egorov S, Daraselia N, et al. Pathway studio—the analysis and navigation of molecular networks. Bioinformatics 2003; 19:2155–2157
5. Wei C-H, Kao H-Y, Lu Z. PubTator: a web-based text mining tool for assisting biocuration. Nucleic Acids Res. 2013; 41:W518-522
6. Rinaldi F, Clematide S, Marques H, et al. OntoGene web services for biomedical text mining. BMC Bioinformatics 2014; 15 Suppl 14:S6
7. Wiegers TC, Davis AP, Mattingly CJ. Web services-based text-mining demonstrates broad impacts for interoperability and process simplification. Database 2014; 2014:1–16
8. Harpaz R, Callahan A, Tamang S, et al. Text mining for adverse drug events: the promise, challenges, and state of the art. Drug Saf 2014; 37:777–790
9. Smith L, Tanabe LK, Ando RJ, et al. Overview of BioCreative II gene mention recognition. Genome biology 2008; 9:S2
10. Uzuner è„°zlem, South BR, Shen S, et al. 2010 i2b2/VA challenge on concepts, assertions, and relations in clinical text. Journal of the American Medical Informatics Association 2011; 18:552–556
11. Krallinger M, Leitner F, Rabal O, et al. Overview of the chemical compound and drug name recognition (CHEMDNER) task. BioCreative Challenge Evaluation Workshop 2013; 2:2–33
12. Ananiadou S, Pyysalo S, Tsujii J 'ichi, et al. Event extraction for systems biology by text mining the literature. Trends Biotechnol. 2010; 28:381–390

13. Van Landeghem S, Björne J, Wei C-H, et al. Large-scale event extraction from literature with multi-level gene normalization. PloS one 2013; 8:e55814
14. Nédellec C, Bossy R, Kim J-D, et al. Overview of bionlp shared task 2013. Proceedings of the BioNLP Shared Task 2013 Workshop 2013; 1–7
15. Krallinger M, Vazquez M, Leitner F, et al. The Protein-Protein Interaction tasks of BioCreative III: classification/ranking of articles and linking bio-ontology concepts to full text. BMC bioinformatics 2011; 12:S3
16. Segura-Bedmar I, Martınez P, Herrero-Zazo M. Semeval-2013 task 9: Extraction of drug-drug interactions from biomedical texts (DDIExtraction 2013). Seventh International Workshop on Semantic Evaluation (SemEval 2013) 2013; 2:341–350
17. Shetty KD, Dalal SR. Using information mining of the medical literature to improve drug safety. J Am Med Inform Assoc 2011; 18:668–674
18. Li C, Liakata M, Rebholz-Schuhmann D. Biological network extraction from scientific literature: state of the art and challenges. Briefings in bioinformatics 2013; 10.1093/bib/bbt006: 1-22
19. Meystre SM, Savova GK, Kipper-Schuler KC, et al. Extracting information from textual documents in the electronic health record: a review of recent research. Yearb Med Inform 2008; 35:128–44
20. Sarker A, Ginn R, Nikfarjam A, et al. Utilizing social media data for pharmacovigilance: A review. Journal of Biomedical Informatics 2015; 54:202–212
21. Piskorski J, Yangarber R. Information extraction: Past, present and future. Multi-source, Multilingual Information Extraction and Summarization 2013; 23–49
22. Simpson MS, Demner-Fushman D. Biomedical text mining: A survey of recent progress. Mining text data 2012; 465–517
23. Zhou D, Zhong D, He Y. Biomedical relation extraction: from binary to complex. Comput Math Methods Med 2014; 2014:298473:1-18
24. Cohen AM, Hersh WR. A survey of current work in biomedical text mining. Briefings in bioinformatics 2005; 6:57–71
25. Zweigenbaum P, Demner-Fushman D, Yu H, et al. Frontiers of biomedical text mining: current progress. Briefings in bioinformatics 2007; 8:358–375
26. Krallinger M, Valencia A, Hirschman L. Linking genes to literature: text mining, information extraction, and retrieval applications for biology. Genome biology 2008; 9:1–14
27. Chapman WW, Cohen KB. Current issues in biomedical text mining and natural language processing. Journal of biomedical informatics 2009; 42:757–759
28. Grishman R. Information extraction: Capabilities and challenges. Notes prepared for the 2012;
29. Hahn U, Cohen KB, Garten Y, et al. Mining the pharmacogenomics literature—a survey of the state of the art. Briefings in bioinformatics 2012; 13:460–494
30. Jiang J. Information extraction from text. Mining text data 2012; 11–41
31. Cohen KB, Demner-Fushman D. Biomedical natural language processing. 2014; 11:
32. Ananiadou S, Thompson P, Nawaz R, et al. Event-based text mining for biology and functional genomics. Brief Funct Genomics 2015; 14:213–230
33. Gonzalez GH, Tahsin T, Goodale BC, et al. Recent Advances and Emerging Applications in Text and Data Mining for Biomedical Discovery. Brief. Bioinformatics 2016; 17:33–42
34. Tang B, Wu Y, Jiang M, et al. Recognizing and Encoding Disorder Concepts in Clinical Text using Machine Learning and Vector Space Model. Working Notes for CLEF 2013 Conference 2013; 1179:
35. Collobert R, Weston J, Bottou L, et al. Natural language processing (almost) from scratch. The Journal of Machine Learning Research 2011; 12:2493–2537
36. de Bruijn B, Cherry C, Kiritchenko S, et al. Machine-learned solutions for three stages of clinical information extraction: the state of the art at i2b2 2010. J Am Med Inform Assoc 2011; 18:557–562

37. Tang B, Cao H, Wu Y, et al. Recognizing clinical entities in hospital discharge summaries using Structural Support Vector Machines with word representation features. BMC Med Inform Decis Mak 2013; 13 Suppl 1:S1
38. Leaman R, Wei C-H, Lu Z. NCBI at the BioCreative IV CHEMDNER Task: Recognizing chemical names in PubMed articles with tmChem. BioCreative Challenge Evaluation Workshop 2013; 2:34–41
39. Xu Y, Hong K, Tsujii J, et al. Feature engineering combined with machine learning and rule-based methods for structured information extraction from narrative clinical discharge summaries. J Am Med Inform Assoc 2012; 19:824–832
40. Miwa M, Sætre R, Kim J-D, et al. Event extraction with complex event classification using rich features. Journal of bioinformatics and computational biology 2010; 8:131–146
41. Björne J, Ginter F, Salakoski T. University of Turku in the BioNLP'11 Shared Task. BMC bioinformatics 2012; 13:S4
42. Rink B, Harabagiu S, Roberts K. Automatic extraction of relations between medical concepts in clinical texts. Journal of the American Medical Informatics Association 2011; 18:594–600
43. Leaman R, Wei C-H, Lu Z. tmChem: a high performance approach for chemical named entity recognition and normalization. J Cheminform 2015; 7:S3
44. Seol J-W, Jo S-H, Yi W, et al. A Problem-Action Relation Extraction Based on Causality Patterns of Clinical Events in Discharge Summaries. Proceedings of the 23rd ACM International Conference on Conference on Information and Knowledge Management(CIKM) 2014; 1971–1974
45. Bui Q-C, Sloot PMA, van Mulligen EM, et al. A novel feature-based approach to extract drug-drug interactions from biomedical text. Bioinformatics 2014; 30:3365–3371
46. Tang B, Cao H, Wang X, et al. Evaluating word representation features in biomedical named entity recognition tasks. Biomed Res Int 2014; 2014:240403:1-6
47. Zheng J, Yarzebski J, Ramesh BP, et al. Automatically Detecting Acute Myocardial Infarction Events from EHR Text: A Preliminary Study Jiaping Zheng, Jorge Yarzebski, Balaji Polepalli Ramesh, Robert J. Goldberg, Hong Yu Proc of AMIA Annual Symposium, 2014, 1286--1293. Proceedings of AMIA Annual Symposium 2014; 1286--1293
48. Tang B, Cao H, Wu Y, et al. Clinical entity recognition using structural support vector machines with rich features. Proceedings of the ACM sixth international workshop on Data and text mining in biomedical informatics 2012; 13–20
49. Minard A-L, Ligozat A-L, Grau B. Multi-class SVM for Relation Extraction from Clinical Reports. RANLP 2011; 604–609
50. Leaman JR. Advancing Biomedical Named Entity Recognition with Multivariate Feature Selection and Semantically Motivated Features. Arizona State University Ph.D. Dissertation 2013; 1–120
51. Xia J, Fang AC, Zhang X, et al. A Novel Feature Selection Strategy for Enhanced Biomedical Event Extraction Using the Turku System. BioMed Research International 2014; 2014:e205239:1-12
52. Campos D, Bui Q-C, Matos S, et al. TrigNER: automatically optimized biomedical event trigger recognition on scientific documents. Source Code for Biology and Medicine 2014; 9:1
53. Fang L, Zhao H, Wang P, et al. Feature selection method based on mutual information and class separability for dimension reduction in multidimensional time series for clinical data. Biomedical Signal Processing and Control 2015; 21:82–89
54. Bui Q-C, Katrenko S, Sloot PM. A hybrid approach to extract protein-protein interactions. Bioinformatics 2011; 27:259–265
55. Patra R, Saha SK. A kernel-based approach for biomedical named entity recognition. ScientificWorldJournal 2013; 2013:950796
56. Zhang Y, Lin H, Yang Z, et al. Hash subgraph pairwise kernel for protein-protein interaction extraction. IEEE/ACM Trans Comput Biol Bioinform 2012; 9:1190–1202

57. Thomas P, Neves M, Solt I, et al. Relation extraction for drug-drug interactions using ensemble learning. 1st Challenge task on Drug-Drug Interaction Extraction (DDIExtraction 2011) 2011; 11–18

58. Yang Z, Tang N, Zhang X, et al. Multiple kernel learning in protein-protein interaction extraction from biomedical literature. Artif Intell Med 2011; 51:163–173

59. Li L, Zhang P, Zheng T, et al. Integrating semantic information into multiple kernels for protein-protein interaction extraction from biomedical literatures. PLoS ONE 2014; 9:e91898

60. Chowdhury MFM, Lavelli A. FBK-irst: A multi-phase kernel based approach for drug-drug interaction detection and classification that exploits linguistic information. Atlanta, Georgia, USA 2013; 351:53

61. Liu J, Xiao L, Shao X. A new approach to extract biomedical events based on composite kernel. Information Science and Technology (ICIST), 2013 International Conference on 2013; 39–42

62. Tikk D, Solt I, Thomas P, et al. A detailed error analysis of 13 kernel methods for protein-protein interaction extraction. BMC Bioinformatics 2013; 14:12

63. Zhou D, Zhong D, He Y. Event trigger identification for biomedical events extraction using domain knowledge. Bioinformatics 2014; 30:1587–1594

64. Li L, Guo R, Jiang Z, et al. An approach to improve kernel-based Protein-Protein Interaction extraction by learning from large-scale network data. Methods 2015; 83:44–50

65. Ma C, Zhang Y, Zhang M. Tree Kernel-based Protein-Protein Interaction Extraction Considering Both Modal Verb Phrases and Appositive Dependency Features. Proceedings of the 24th International Conference on World Wide Web 2015; 655–660

66. Kim S, Liu H, Yeganova L, et al. Extracting drug-drug interactions from literature using a rich feature-based linear kernel approach. J Biomed Inform 2015; 55:23–30

67. Lingren T, Deleger L, Molnar K, et al. Evaluating the impact of pre-annotation on annotation speed and potential bias: natural language processing gold standard development for clinical named entity recognition in clinical trial announcements. J Am Med Inform Assoc 2014; 21:406–413

68. South BR, Mowery D, Suo Y, et al. Evaluating the effects of machine pre-annotation and an interactive annotation interface on manual de-identification of clinical text. Journal of Biomedical Informatics 2014; 50:162–172

69. Gobbel GT, Garvin J, Reeves R, et al. Assisted annotation of medical free text using RapTAT. J Am Med Inform Assoc 2014; 21:833–841

70. Chen Y, Mani S, Xu H. Applying active learning to assertion classification of concepts in clinical text. J Biomed Inform 2012; 45:265–272

71. Chen Y, Cao H, Mei Q, et al. Applying active learning to supervised word sense disambiguation in MEDLINE. Journal of the American Medical Informatics Association 2013; 20:1001–1006

72. Chen Y, Lasko TA, Mei Q, et al. A study of active learning methods for named entity recognition in clinical text. J Biomed Inform 2015; 58:11–18

73. Miller TA, Dligach D, Savova GK. Active Learning for Coreference Resolution. Proceedings of the 2012 Workshop on Biomedical Natural Language Processing 2012; 73–81

74. Hahn U, Beisswanger E, Buyko E, et al. Active Learning-Based Corpus Annotation—The PathoJen Experience. AMIA Annu Symp Proc 2012; 2012:301–310

75. Dligach D, Miller TA, Savova GK. Active Learning for Phenotyping Tasks. Proceedings of the Workshop on NLP for Medicine and Biology associated with RANLP 2013; 1–8

76. Kholghi M, Sitbon L, Zuccon G, et al. External Knowledge and Query Strategies in Active Learning: A Study in Clinical Information Extraction. Proceedings of the 24th ACM International on Conference on Information and Knowledge Management 2015; 143–152

77. Kholghi M, Sitbon L, Zuccon G, et al. Active learning: a step towards automating medical concept extraction. J Am Med Inform Assoc 2016; 23:289–296

78. Khare R, Good BM, Leaman R, et al. Crowdsourcing in biomedicine: challenges and opportunities. Brief. Bioinformatics 2016; 17:23–32

79. Zhai H, Lingren T, Deleger L, et al. Web 2.0-Based Crowdsourcing for High-Quality Gold Standard Development in Clinical Natural Language Processing. Journal of Medical Internet Research 2013; 15:e73
80. Good BM, Nanis M, Wu C, et al. Microtask crowdsourcing for disease mention annotation in PubMed abstracts. Pac Symp Biocomput 2015; 282–293
81. McCoy AB, Wright A, Laxmisan A, et al. Development and evaluation of a crowdsourcing methodology for knowledge base construction: identifying relationships between clinical problems and medications. J Am Med Inform Assoc 2012; 19:713–718
82. Demartini G, Difallah DE, Cudré-Mauroux P. ZenCrowd: Leveraging Probabilistic Reasoning and Crowdsourcing Techniques for Large-scale Entity Linking. Proceedings of the 21st International Conference on World Wide Web 2012; 469–478
83. Dumitrache A, Aroyo L, Welty C, et al. 'Dr. Detective': Combining Gamification Techniques and Crowdsourcing to Create a Gold Standard in Medical Text. Proceedings of the 1st International Conference on Crowdsourcing the Semantic Web - Volume 1030 2013; 16–31
84. Dumitrache A, Aroyo L, Welty C. Achieving Expert-Level Annotation Quality with CrowdTruth. Proc. of BDM2I Workshop, ISWC, 2015
85. Zhang S, Elhadad N. Unsupervised biomedical named entity recognition: experiments with clinical and biological texts. J Biomed Inform 2013; 46:1088–1098
86. Quan C, Wang M, Ren F. An unsupervised text mining method for relation extraction from biomedical literature. PLoS ONE 2014; 9:e102039
87. Alicante A, Corazza A, Isgrò F, et al. Unsupervised entity and relation extraction from clinical records in Italian. Computers in Biology and Medicine 2016; 263–275
88. Jochim C, Sacaleanu B, Deleris LA. Risk Event and Probability Extraction for Modeling Medical Risks. 2014 AAAI Fall Symposium Series 2014;
89. Xu R, Wang Q. A semi-supervised approach to extract pharmacogenomics-specific drug-gene pairs from biomedical literature for personalized medicine. J Biomed Inform 2013; 46:585–593
90. Lv X, Guan Y, Deng B. Transfer learning based clinical concept extraction on data from multiple sources. J Biomed Inform 2014; 52:55–64
91. Wang C, Fan J. Medical Relation Extraction with Manifold Models. ACL (1) 2014; 828–838
92. Song M, Yu H, Han W-S. Combining active learning and semi-supervised learning techniques to extract protein interaction sentences. BMC Bioinformatics 2011; 12:1–11
93. Li P, Zhu Q, Zhou G. Employing Event Inference to Improve Semi-Supervised Chinese Event Extraction. COLING 2014; 2161–2171
94. Zhou D, Zhong D. A semi-supervised learning framework for biomedical event extraction based on hidden topics. Artif Intell Med 2015; 64:51–58
95. Li Y, Hu X, Lin H, et al. A framework for semisupervised feature generation and its applications in biomedical literature mining. IEEE/ACM Transactions on Computational Biology and Bioinformatics (TCBB) 2011; 8:294–307
96. Wang J, Xu Q, Lin H, et al. Semi-supervised method for biomedical event extraction. Proteome Sci 2013; 11:1–10
97. Thomas P, Bobić T, Leser U, et al. Weakly Labeled Corpora as Silver Standard for Drug-Drug and Protein-Protein Interaction. Proceedings of the Workshop on Building and Evaluating Resources for Biomedical Text Mining (BioTxtM) on Language Resources and Evaluation Conference (LREC) 2012; 63–70
98. Provoost T, Moens M-F. Semi-supervised Learning for the BioNLP Gene Regulation Network. BMC Bioinformatics 2015; 16:S4
99. Bobić T, Klinger R, Thomas P, et al. Improving distantly supervised extraction of drug-drug and protein-protein interactions. Proceedings of the Joint Workshop on Unsupervised and Semi-Supervised Learning in NLP 2012; 35–43

100. Bobic T, Klinger R. Committee-based Selection of Weakly Labeled Instances for Learning Relation Extraction. Proceedings of the Conference on Intelligent Text Processing and Computational Linguistics 2013; 70:112

101. Liu M, Ling Y, An Y, et al. Relation extraction from biomedical literature with minimal supervision and grouping strategy. 2014 IEEE International Conference on Bioinformatics and Biomedicine (BIBM) 2014; 444–449

102. Poon H, Toutanova K, Quirk C. Distant supervision for cancer pathway extraction from text. Pac Symp Biocomput 2015; 120–131

103. Mallory EK, Zhang C, Ré C, et al. Large-scale extraction of gene interactions from full-text literature using DeepDive. Bioinformatics 2016; 32:106–113

104. WEI X, ZHU Q, LYU C, et al. A Hybrid Method to Extract Triggers in Biomedical Events. Journal of Digital Information Management 2015; 13:299

105. Javed R, Farhan S, Humdullah S. A Hybrid Approach Based on Pattern Recognition and BioNLP for Investigating Drug-Drug Interaction. Current Bioinformatics 2015; 10:315–322

106. Chang Y-C, Dai H-J, Wu JC-Y, et al. TEMPTING system: a hybrid method of rule and machine learning for temporal relation extraction in patient discharge summaries. J Biomed Inform 2013; 46 Suppl:S54-62

107. Wang C, Akella R. A Hybrid Approach to Extracting Disorder Mentions from Clinical Notes. AMIA Jt Summits Transl Sci Proc 2015; 2015:183–187

108. Muzaffar AW, Azam F, Qamar U. A Relation Extraction Framework for Biomedical Text Using Hybrid Feature Set. Computational and Mathematical Methods in Medicine 2015; 2015:1–12

109. Roller R, Stevenson M. Identification of Genia Events using Multiple Classifiers. Proceedings of BioNLP Shared Task Workshop 2013; 125–129

110. Wei C-H, Leaman R, Lu Z. SimConcept: A Hybrid Approach for Simplifying Composite Named Entities in Biomedical Text. IEEE Journal of Biomedical and Health Informatics 2015; 19:1385–1391

111. Poon H, Vanderwende L. Joint inference for knowledge extraction from biomedical literature. Human Language Technologies: The 2010 Annual Conference of the North American Chapter of the Association for Computational Linguistics 2010; 813–821

112. Riedel S, McCallum A. Fast and robust joint models for biomedical event extraction. Proceedings of the Conference on Empirical Methods in Natural Language Processing 2011; 1–12

113. Vlachos A, Craven M. Biomedical event extraction from abstracts and full papers using search-based structured prediction. BMC Bioinformatics 2012; 13 Suppl 11:S5

114. Venugopal D, Chen C, Gogate V, et al. Relieving the Computational Bottleneck: Joint Inference for Event Extraction with High-Dimensional Features. EMNLP 2014; 831–843

115. Li L, Liu S, Qin M, et al. Extracting Biomedical Event with Dual Decomposition Integrating Word Embeddings. IEEE/ACM Trans Comput Biol Bioinform 2015; PP:1–1

116. Wei X, Huang Y, Lyu C, et al. Extracting Nested Biomedical Entity Relations by Tagging Dependency Chains. Journal of Engineering Science & Technology Review 2015; 8:50

117. Li F, Ji D, Wei X, et al. A transition-based model for jointly extracting drugs, diseases and adverse drug events. 2015 IEEE International Conference on Bioinformatics and Biomedicine (BIBM) 2015; 599–602

118. Banko M, Cafarella MJ, Soderland S, et al. Open information extraction for the web. IJCAI 2007; 7:2670–2676

119. Wu F, Weld DS. Open Information Extraction Using Wikipedia. Proceedings of the 48th Annual Meeting of the Association for Computational Linguistics 2010; 118–127

120. Fader A, Soderland S, Etzioni O. Identifying Relations for Open Information Extraction. Proceedings of the Conference on Empirical Methods in Natural Language Processing 2011; 1535–1545

121. Etzioni O, Fader A, Christensen J, et al. Open Information Extraction: The Second Generation. IJCAI 2011; 11:3–10

122. Akbik A, Löser A. Kraken: N-ary facts in open information extraction. Proceedings of the Joint Workshop on Automatic Knowledge Base Construction and Web-scale Knowledge Extraction 2012; 52–56
123. Mausam, Schmitz M, Bart R, et al. Open Language Learning for Information Extraction. Proceedings of the 2012 Joint Conference on Empirical Methods in Natural Language Processing and Computational Natural Language Learning 2012; 523–534
124. Pujara J, Getoor L. Building Dynamic Knowledge Graphs. NIPS Workshop on Automated Knowledge Base Construction 2014;
125. Dong X, Gabrilovich E, Heitz G, et al. Knowledge vault: A web-scale approach to probabilistic knowledge fusion. Proceedings of the 20th ACM SIGKDD international conference on Knowledge discovery and data mining 2014; 601–610
126. Nakashole N, Weikum G, Suchanek F. PATTY: A Taxonomy of Relational Patterns with Semantic Types. Proceedings of the 2012 Joint Conference on Empirical Methods in Natural Language Processing and Computational Natural Language Learning 2012; 1135–1145
127. Callahan A, Dumontier M, Shah NH. HyQue: evaluating hypotheses using Semantic Web technologies. J Biomed Semantics 2011; 2:S3
128. Soderland S, Roof B, Qin B, et al. Adapting open information extraction to domain-specific relations. AI Magazine 2010; 31:93–102
129. Lin T, Mausam, Etzioni O. No Noun Phrase Left Behind: Detecting and Typing Unlinkable Entities. Proceedings of the 2012 Joint Conference on Empirical Methods in Natural Language Processing and Computational Natural Language Learning 2012; 893–903
130. Wijaya D, Talukdar PP, Mitchell T. PIDGIN: Ontology Alignment Using Web Text As Interlingua. Proceedings of the 22Nd ACM International Conference on Conference on Information & Knowledge Management 2013; 589–598
131. Riedel S, Yao L, McCallum A, et al. Relation extraction with matrix factorization and universal schemas. Proceedings of NAACL-HLT 2013; 74–84
132. Movshovitz-Attias D, Cohen WW. Bootstrapping biomedical ontologies for scientific text using nell. Proceedings of the 2012 Workshop on Biomedical Natural Language Processing 2012; 11–19
133. Carlson A, Betteridge J, Kisiel B, et al. Toward an Architecture for Never-Ending Language Learning. AAAI 2010; 5:3–10
134. Nebot V, Berlanga R. Semantics-aware open information extraction in the biomedical domain. Proceedings of the 4th International Workshop on Semantic Web Applications and Tools for the Life Sciences 2011; 84–91
135. Nebot V, Berlanga R. Exploiting Semantic Annotations for Open Information Extraction: an experience in the biomedical domain. Knowledge and information Systems 2014; 38:365–389
136. Deng L, Yu D. Deep Learning: Methods and Applications. Foundations and Trends in Signal Processing 2014; 7:197–387
137. Salakhutdinov R, Mnih A, Hinton G. Restricted Boltzmann Machines for Collaborative Filtering. Proceedings of the 24th International Conference on Machine Learning 2007; 791–798
138. Hinton G, Osindero S, Teh Y-W. A fast learning algorithm for deep belief nets. Neural computation 2006; 18:1527–1554
139. Lamblin P, Bengio Y. Important gains from supervised fine-tuning of deep architectures on large labeled sets. NIPS* 2010 Deep Learning and Unsupervised Feature Learning Workshop 2010;
140. Krizhevsky A, Sutskever I, Hinton GE. Imagenet classification with deep convolutional neural networks. Advances in neural information processing systems 2012; 1097–1105
141. Socher R, Lin CC, Manning C, et al. Parsing natural scenes and natural language with recursive neural networks. Proceedings of the 28th international conference on machine learning (ICML-11) 2011; 129–136

142. Erhan D, Bengio Y, Courville A, et al. Why does unsupervised pre-training help deep learning? The Journal of Machine Learning Research 2010; 11:625–660
143. Yih W, He X, Meek C. Semantic Parsing for Single-Relation Question Answering. ACL (2) 2014; 643–648
144. Shen Y, He X, Gao J, et al. Learning Semantic Representations Using Convolutional Neural Networks for Web Search. Proceedings of the Companion Publication of the 23rd International Conference on World Wide Web Companion 2014; 373–374
145. Severyn A, Moschitti A. Learning to Rank Short Text Pairs with Convolutional Deep Neural Networks. Proceedings of the 38th International ACM SIGIR Conference on Research and Development in Information Retrieval 2015; 373–382
146. Zhou J, Xu W. End-to-end learning of semantic role labeling using recurrent neural networks. Proceedings of the Annual Meeting of the Association for Computational Linguistics 2015; 1127–1137
147. Mazalov A, Martins B, Matos D. Spatial role labeling with convolutional neural networks. Proceedings of the 9th Workshop on Geographic Information Retrieval 2015; 12
148. Severyn A, Moschitti A. Twitter sentiment analysis with deep convolutional neural networks. Proceedings of the 38th International ACM SIGIR Conference on Research and Development in Information Retrieval 2015; 959–962
149. Iyyer M, Boyd-Graber J, Claudino L, et al. A Neural Network for Factoid Question Answering over Paragraphs. Empirical Methods in Natural Language Processing 2014; 633–644
150. Yu L, Hermann KM, Blunsom P, et al. Deep Learning for Answer Sentence Selection. NIPS Deep Learning and Representation Learning Workshop 2014;
151. Kumar A, Irsoy O, Ondruska P, et al. Ask Me Anything: Dynamic Memory Networks for Natural Language Processing. NIPS Deep Learning Symposium 2015;
152. Yin W, Ebert S, Schütze H. Attention-Based Convolutional Neural Network for Machine Comprehension. arXiv preprint arXiv:1602.04341 2016;
153. Cho K, van Merrienboer B, Gulcehre C, et al. Learning phrase representations using rnn encoder-decoder for statistical machine translation. arXiv preprint arXiv:1406.1078 2014;
154. Sutskever I, Vinyals O, Le QV. Sequence to sequence learning with neural networks. Advances in Neural Information Processing Systems 2014; 3104–3112
155. Luong M-T, Le QV, Sutskever llya, et al. Multi-task Sequency to Sequence Learning. Proceedings of ICLR 2016
156. Firat O, Cho K, Bengio Y. Multi-Way, Multilingual Neural Machine Translation with a Shared Attention Mechanism. arXiv preprint arXiv:1601.01073 2016;
157. Feng S, Liu S, Li M, et al. Implicit Distortion and Fertility Models for Attention-based Encoder-Decoder NMT Model. arXiv preprint arXiv:1601.03317 2016;
158. Liu P, Qiu X, Chen X, et al. Multi-Timescale Long Short-Term Memory Neural Network for Modelling Sentences and Documents. Proceedings of the 2015 Conference on Empirical Methods in Natural Language Processing 2326–2335
159. Wu H, Gu Y, Sun S, et al. Aspect-based Opinion Summarization with Convolutional Neural Networks. arXiv preprint arXiv:1511.09128 2015;
160. Marujo L, Ling W, Ribeiro R, et al. Exploring events and distributed representations of text in multi-document summarization. Knowledge-Based Systems 2016; 94:33–42
161. Graves A. Generating Sequences With Recurrent Neural Networks. arXiv preprint arXiv:1308.0850 2013;
162. Huang H, Heck L, Ji H. Leveraging Deep Neural Networks and Knowledge Graphs for Entity Disambiguation. arXiv preprint arXiv:1504.07678 2015;
163. Nguyen TH, Sil A, Dinu G, et al. Toward Mention Detection Robustness with Recurrent Neural Networks. arXiv preprint arXiv:1602.07749 2016;

164. Nguyen TH, Grishman R. Combining Neural Networks and Log-linear Models to Improve Relation Extraction. arXiv preprint arXiv:1511.05926 2015;
165. Yan X, Mou L, Li G, et al. Classifying Relations via Long Short Term Memory Networks along Shortest Dependency Path. Proceedings of the 2015 Conference on Empirical Methods in Natural Language Process 2015; 1785–1794
166. Miwa M, Bansal M. End-to-end Relation Extraction using LSTMs on Sequences and Tree Structures. Proceedings of ACL 2016;
167. Xu Y, Jia R, Mou L, et al. Improved Relation Classification by Deep Recurrent Neural Networks with Data Augmentation. arXiv preprint arXiv:1601.03651 2016;
168. Qin P, Xu W, Guo J. An Empirical Convolutional Neural Network approach for Semantic Relation Classification. Neurocomputing 2016; 1–9
169. Dasigi P, Hovy EH. Modeling Newswire Events using Neural Networks for Anomaly Detection. Proceedings of COLING 2014, the 25th International Conference on Computational Linguists 2014; 1414–1422
170. Nguyen TH, Grishman R. Event detection and domain adaptation with convolutional neural networks. Proceedings of ACL 2015; 365–371
171. Chen Y, Xu L, Liu K, et al. Event Extraction via Dynamic Multi-Pooling Convolutional Neural Networks. Proceedings of ACL 2015; 167–176
172. Liu S, Tang B, Chen Q, et al. Effects of Semantic Features on Machine Learning-Based Drug Name Recognition Systems: Word Embeddings vs. Manually Constructed Dictionaries. Information 2015; 6:848–865
173. Jagannatha AN, Chen J, Yu H. Mining and Ranking Biomedical Synonym Candidates from Wikipedia. Proceedings of the Sixth International Workshop on Health Text Mining and Information Analysis (Louhi) 2015; 142–151
174. De Vine L, Zuccon G, Koopman B, et al. Medical semantic similarity with a neural language model. Proceedings of the 23rd ACM International Conference on Conference on Information and Knowledge Management 2014; 1819–1822
175. Jiang Z, Jin L, Li L, et al. A CRD-WEL System for Chemical-disease Relations Extraction. Proceedings of the Fifth BioCreative Challenge Evaluation Workshop 2015; 317–326
176. Jiang Z, Li shuang, Huang D. A general protein-protein interaction extraction architecture based on word representation and feature selection. International Journal of Data Mining and Bioinformatics 2016; 14:276–291
177. Li C, Song R, Liakata M, et al. Using word embedding for bio-event extraction. Proceedings of the 2015 Workshop on Biomedical Natural Language Processing 2015; 121–126
178. Nie Y, Rong W, Zhang Y, et al. Embedding assisted prediction architecture for event trigger identification. Journal of bioinformatics and computational biology 2015; 13:1541001
179. Henriksson A, Kvist M, Dalianis H, et al. Identifying adverse drug event information in clinical notes with distributional semantic representations of context. Journal of biomedical informatics 2015; 57:333–349
180. Wu Y, Xu J, Zhang Y, et al. Clinical Abbreviation Disambiguation Using Neural Word Embeddings. Proceedings of the 2015 Workshop on Biomedical Natural Language Processing 2015; 171–176
181. Liu Y, Ge T, Mathews KS, et al. Exploiting Task-Oriented Resources to Learn Word Embeddings for Clinical Abbreviation Expansion. Proceedings of the 2015 Workshop on Biomedical Natural Language Processing 2015; 92–97
182. Ghassemi MM, Mark RG, Nemati S. Visualizing Evolving Clinical Sentiment Using Vector Representations of Clinical Notes and distributed stochastic neighbor embedding. Computing in Cardiology Conference (CinC) 2015; 629–632

183. Jiang Z, Li L, Huang D, et al. Training word embeddings for deep learning in biomedical text mining tasks. 2015 IEEE International Conference on Bioinformatics and Biomedicine (BIBM) 2015; 625–628

184. Choi E, Bahadori MT, Searles E, et al. Multi-layer Representation Learning for Medical Concepts. Proceedings of 22nd ACM SIGKDD Conference on Knowledge Discovery and Data Mining 2016;

185. Choi Y. Learning Low-Dimensional Representations of Medical Concepts. Proceedings of the AMIA Summit on Clinical Research Informatics (CRI) 2016;

186. Choi E, Bahadori MT, Sun J. Doctor AI: Predicting Clinical Events via Recurrent Neural Networks. arXiv:1511.05942 [cs] 2015;

187. Hochreiter S, Schmidhuber J. Long short-term memory. Neural computation 1997; 9:1735–1780

188. Abhyuday Jagannatha, Hong Yu. Bidirectional RNN for Medical Event Detection in Electronic Health Records. NAACL HLT 2016. 2016;

189. Miotto R, Li L, Kidd BA, et al. Deep Patient: An Unsupervised Representation to Predict the Future of Patients from the Electronic Health Records. Sci Rep 2016; 6:

190. Liu Y-T, Lin Y-Y, Wu S-L, et al. Brain Dynamics in Predicting Driving Fatigue Using a Recurrent Self-Evolving Fuzzy Neural Network. IEEE Transactions on Neural Networks and Learning Systems 2016; 27:347–360

191. Lehman L, Ghassemi MM, Snoek J, et al. Patient Prognosis from Vital Sign Time Series: Combining Convolutional Neural Networks with a Dynamical Systems Approach. In Computing in Cardiology Conference (Cinc) 2015;

192. Ginneken B van, Setio AAA, Jacobs C, et al. Off-the-shelf convolutional neural network features for pulmonary nodule detection in computed tomography scans. 2015 IEEE 12th International Symposium on Biomedical Imaging (ISBI) 2015; 286–289

193. Bar Y, Diamant I, Wolf L, et al. Chest pathology detection using deep learning with non-medical training. 2015 IEEE 12th International Symposium on Biomedical Imaging (ISBI) 2015; 294–297

194. Roth HR, Wang Y, Yao J, et al. Deep convolutional networks for automated detection of posterior-element fractures on spine CT. arXiv:1602.00020 [cs] 2016;

195. Shin H-C, Lu L, Kim L, et al. Interleaved text/image Deep Mining on a large-scale radiology database. 2015 IEEE Conference on Computer Vision and Pattern Recognition (CVPR) 2015; 1090–1099

196. Shin HC, Roth HR, Gao M, et al. Deep Convolutional Neural Networks for Computer-Aided Detection: CNN Architectures, Dataset Characteristics and Transfer Learning. IEEE Transactions on Medical Imaging 2016; PP:1–1

197. Zhang S, Zhou J, Hu H, et al. A deep learning framework for modeling structural features of RNA-binding protein targets. Nucleic acids research 2015; gkv1025

198. Liu F, Ren C, Li H, et al. *De novo* identification of replication-timing domains in the human genome by deep learning. Bioinformatics 2016; 32:641–649

199. Leung MK, Delong A, Babak Alipanahi, et al. Machine Learning in Genomic Medicine: A Review of Computational Problems and Data Sets. Proceedings of IEEE 2016; 104:176–197

200. Spencer M, Eickholt J, Cheng J. A Deep Learning Network Approach to ab initio Protein Secondary Structure Prediction. IEEE/ACM Transactions on Computational Biology and Bioinformatics 2015; 12:103–112

201. Gawehn E, Hiss JA, Schneider G. Deep Learning in Drug Discovery. Molecular Informatics 2016; 35:3–14

202. Toutanova K, Chen D, Pantel P, et al. Representing text for joint embedding of text and knowledge bases. Proceedings of the 2015 Conference on Empirical Methods in Natural Language Processing 2015; 1499–1509

203. Guo S, Wang Q, Wang B, et al. Semantically smooth knowledge graph embedding. Proceedings of the 53rd Annual Meeting of the Association for Computational Linguistics and the 7th International Joint Conference on Natural Language Processing 2015; 84–94

204. Xiao H, Huang M, Hao Y, et al. TransA: An Adaptive Approach for Knowledge Graph Embedding. arXiv preprint arXiv:1509.05490 2015;
205. Luo Y, Wang Q, Wang B, et al. Context-Dependent Knowledge Graph Embedding. Proceedings of the 2015 Conference on Empirical Methods in Natural Language Processing 2015; 1656–1661
206. Ding X. BUEES: a bottom-up event extraction system. Frontiers of Information Technology & Electronic Engineering 2015; 16:541–552
207. Riedel S, Chun H-W, Takagi T, et al. A markov logic approach to bio-molecular event extraction. Proceedings of the Workshop on Current Trends in Biomedical Natural Language Processing: Shared Task 2009; 41–49
208. Poon H, Vanderwende L. Joint inference for knowledge extraction from biomedical literature. Human Language Technologies: The 2010 Annual Conference of the North American Chapter of the Association for Computational Linguistics 2010; 813–821
209. Collobert R, Weston J, Bottou L, et al. Natural language processing (almost) from scratch. The Journal of Machine Learning Research 2011; 12:2493–2537
210. Bordes A, Weston J, Collobert R, et al. Learning structured embeddings of knowledge bases. Conference on Artificial Intelligence 2011;
211. Bordes A, Glorot X, Weston J, et al. A semantic matching energy function for learning with multi-relational data. Machine Learning 2014; 94:233–259
212. Long M, Cao Y, Wang J, et al. Learning Transferable Features with Deep Adaptation Networks. Proceedings of the 32nd International Conference on Machine Learning (ICML-15) 2015; 97–105
213. Klinger R, Riedel S, McCallum A. Inter-Event Dependencies support Event Extraction from Biomedical Literature. Mining Complex Entities from Network and Biomedical Data (MIND workshop), European Conference on Machine Learning and Principles and Practice of Knowledge Discovery in Databases (ECML PKDD) 2011;
214. Li C, Liakata M. Bio-event definition in text mining towards event interconnection. BMC Proceedings 2015; 9:A5